\documentclass[letterpaper, 10 pt, journal, twoside]{ieeetran}
\usepackage{caption}
\usepackage{amsmath,amsfonts,amssymb}
\usepackage{algorithmic}
\usepackage{algorithm}
\usepackage{array}
\usepackage[caption=false,font=normalsize,labelfont=sf,textfont=sf]{subfig}
\usepackage{textcomp}
\usepackage{stfloats}
\usepackage{url}
\usepackage{verbatim}
\usepackage{graphicx}
\usepackage{cite}
\usepackage{authblk}
\usepackage{xcolor}
\usepackage{wrapfig}
\usepackage{bm}          % bold math symbols
\usepackage{booktabs}
\usepackage{siunitx}
\usepackage{hyperref}
\usepackage{mathtools}

\graphicspath{{figures/}}

\newcommand{\R}{\mathbb{R}}

\newcommand{\SO}{\mathrm{SO}(3)}
\newcommand{\so}{\mathfrak{so}(3)}
\newcommand{\vect}[1]{\bm{#1}}
\newcommand{\mat}[1]{\bm{#1}}

\newcommand{\norm}[1]{\left\lVert #1 \right\rVert}

\newtheorem{remark}{Remark}
\begin{document}

\title{Energy-Efficient Arm Reaching for a Humanoid Robot via Deep
       Reinforcement Learning with Identified Power Models}

\author{Nestor N. Deniz$^{1}$, Simon Parsons$^{2}$ and Fernando A. Auat Cheein$^{1}$
% \thanks{Manuscript received: December, 12, 2024; Revised June, 9, 2025; Accepted August, 11, 2025.}%Use only for final RAL version
\thanks{} %Use only for final RAL version
\thanks{$^{1}$Nestor N. Deniz, Sebastian Vega and Fernando Auat Cheein are with the Engineering Department at Harper Adams University, Newport, Shropshire TF10 8NB, UK.
        {\tt\footnotesize ndeniz@harper-adams.ac.uk}}%
\thanks{$^{2} $Simon Parsons is with Lincoln Institute for Agri-Food Technology and Lincoln Centre for Autonomous Systems.}%
% \thanks{Digital Object Identifier (DOI): see top of this page.}
}

\maketitle

% =============================================================================
\begin{abstract}
Humanoid robots performing in-field manipulation tasks, such as robotic apple
harvesting, face severe energy constraints that directly limit the number of
reaching motions that can be executed per battery charge.  This paper presents
an end-to-end, energy-aware reinforcement learning framework for the
7-degree-of-freedom left arm of the Unitree~G1 humanoid robot, combining a
physics-based, experimentally identified electrical power model with a
Soft Actor-Critic (SAC) policy trained in a Pinocchio-based rigid-body
dynamics simulator.
The RL policy operates on an incremental joint-position
action space and is trained with a Hybrid Constellation Reward that combines a
four-point end-effector constellation distance with a torque-norm energy
proxy; after 
% $5\times10^6$ 
training it reaches a $69.9\%$ success rate
over $1\,000$ random targets in kinematic simulation, at a mean energy of
\SI{98.16}{\joule} on successful episodes. Finally, on the
physical Unitree~G1, the policy is validated over three independent
10-target batches, achieving a mean energy of
$71.5 \pm 48.3$\,J, an end-effector position error of $2.64 \pm 1.04$\,cm, and
an orientation error of $6.92 \pm 1.33^\circ$---within the
\SI{4}{\centi\metre}/$8.6^\circ$ training tolerance.  These results constitute
a first step toward energy-aware reinforcement-learning-based arm reaching for
humanoid robots.
\end{abstract}

% =============================================================================
\section{Introduction}
\label{sec:intro}

\IEEEPARstart{H}{umanoid} robots are increasingly considered for deployment in
field environments---inspection, agriculture, disaster response---where
operation time is dictated by a fixed battery capacity
\cite{Neunert2018,Minniti2019,Pankert2020}.  A particularly demanding instance
of this class of tasks is robotic fruit harvesting: an autonomous apple-picking
humanoid must repeatedly reach toward, align with, and grasp fruit distributed
throughout a tree canopy, executing hundreds of arm-reaching motions per
working session on a single battery charge.  Unlike wheeled platforms, whose
locomotion cost is relatively predictable, the arm joints of a humanoid can
account for a highly variable fraction of total power depending on task speed,
payload, and trajectory shape, and this variability is compounded over the
large number of repetitive reaches required for harvesting.  Reducing arm
energy consumption per reach is therefore critical for extending the
productive working time of a humanoid harvesting platform.

Classical motion-planning methods minimise a surrogate for energy, such as
joint torques~\cite{Siciliano2010} or joint velocities, but they require an
explicit, accurate dynamics model and solve an optimisation problem online,
incurring substantial computational overhead.  Model Predictive Control (MPC)
can incorporate energy objectives within a receding-horizon
framework~\cite{Rawlings2020,Kleff2021}, yet the quality of MPC solutions is
bounded by the fidelity of its internal model, and real-time re-optimisation
at control rates of \SI{100}{\hertz} demands dedicated solvers and often
constrains the problem horizon to only a few steps.

Deep Reinforcement Learning (RL) offers a complementary paradigm: the policy
is trained offline in simulation and executes online as a fixed feedforward
network, adding negligible computation at inference time.  Recent work has
demonstrated RL policies for dexterous manipulation
\cite{OpenAI2019,Andrychowicz2020}, whole-body
locomotion~\cite{Kumar2021,Rudin2022}, and energy-aware quadruped
gaits~\cite{Schulman2017}, but energy-efficient reaching for a physical
humanoid arm---where an experimentally identified power model is embedded
directly into the RL reward, and where the resulting policy is validated
end-to-end from kinematic simulation through dynamics-level simulation to
physical hardware---has received little attention, particularly in the
context of agricultural manipulation.

This paper makes the following contributions:
\begin{enumerate}
  \item An experimentally identified electrical power model for the 7-DOF
        Unitree~G1 left arm, used both as an evaluation metric and as an
        energy proxy inside a Gymnasium RL environment
        (Section~\ref{sec:power_model}).
  \item A Soft Actor-Critic (SAC) \cite{Haarnoja2018} policy, combining an
        incremental joint-position action space with a Hybrid Constellation
        Reward, that achieves a $69.9\%$ success rate over $1\,000$ random
        targets in kinematic simulation (Section~\ref{sec:rl}).
  \item A dynamics-level sim-to-real validation in MuJoCo---including a
        frozen-policy PD-gain sweep of the low-level controller and a
        workspace-reachability analysis---that quantifies and explains the
        gap between kinematic training performance and physically realisable
        performance (Sections~\ref{sec:training} and~\ref{sec:experiments}).
  \item A real-hardware validation campaign on the physical Unitree~G1
        (30 independent reaching trials across three batches), reporting the
        end-effector position and orientation error achieved on every trial
        and showing that the learned policy transfers to a restricted but
        representative apple-picking task envelope
        (Section~\ref{sec:experiments}).
\end{enumerate}

% =============================================================================
\section{Robot Platform and Power Model}
\label{sec:power_model}

\subsection{Unitree G1 Left Arm}
\label{sec:robot}

The Unitree~G1 is a full-size humanoid robot weighing approximately
\SI{35}{\kilo\gram}.  Its left arm consists of seven revolute joints:
shoulder pitch ($q_0$), shoulder roll ($q_1$), shoulder yaw ($q_2$), elbow
($q_3$), wrist roll ($q_4$), wrist pitch ($q_5$), and wrist yaw ($q_6$).
Joint angles, velocities, and motor currents are broadcast via the
Unitree DDS middleware at \SI{500}{\hertz}; main-board power is reported by
an onboard sensor at \SI{1}{\hertz}.  Joint limits, expressed in radians,
are listed in Table~\ref{tab:joint_limits}.

\begin{table}[t]
  \caption{Left-arm joint limits and maximum velocity}
  \label{tab:joint_limits}
  \centering
  \setlength{\tabcolsep}{4pt}
  \begin{tabular}{lcccc}
    \toprule
    Joint & Index & Lower (rad) & Upper (rad) & $\dot{q}_{\max}$ (rad/s) \\
    \midrule
    Shoulder pitch & 0 & $-3.089$ & $ 2.670$ & 2.5 \\
    Shoulder roll  & 1 & $-1.588$ & $ 2.252$ & 2.5 \\
    Shoulder yaw   & 2 & $-2.618$ & $ 2.618$ & 2.5 \\
    Elbow          & 3 & $-1.047$ & $ 2.094$ & 2.5 \\
    Wrist roll     & 4 & $-1.972$ & $ 1.972$ & 2.5 \\
    Wrist pitch    & 5 & $-1.614$ & $ 1.614$ & 2.5 \\
    Wrist yaw      & 6 & $-1.614$ & $ 1.614$ & 2.5 \\
    \bottomrule
  \end{tabular}
\end{table}

\subsection{Physics-Based Power Model}
\label{sec:model_structure}

We model the net electrical power of the arm as a sum of per-joint terms and
pairwise joint-speed interaction terms~\cite{DenizPowerModel2025}:
\begin{equation}
  \label{eq:power}
  \begin{array}{rl}  
  P_{\mathrm{net}}(\vect{\tau},\dot{\vect{q}})
    =& \sum_{i=0}^{n-1}
        \Bigl[
          a_i\,\tau_i\,\dot{q}_i
          + b_i\,\Delta\tau_i^2
          + c_i\,|\dot{q}_i|
          + d_i\,\dot{q}_i^2
        \Bigr]\\
    &+ \sum_{i<j} e_{ij}\,|\dot{q}_i|\,|\dot{q}_j|,
    \end{array}
\end{equation}
where $n=7$, $\vect{\tau}\in\R^7$ is the joint torque vector,
$\dot{\vect{q}}\in\R^7$ is the joint velocity vector, and
\begin{equation}
  \label{eq:delta_tau}
  \Delta\tau_i^2 \triangleq \tau_i^2 - \tau_{0,i}^2
\end{equation}
is the change in copper loss relative to the static baseline torque
$\tau_{0,i}$ required to hold the arm at the neutral configuration ($\vect{q}=\vect{0}$).

The four per-joint terms model, respectively:
\begin{itemize}
  \item \emph{Mechanical power} ($a_i\tau_i\dot{q}_i$): net work done by
        joint~$i$, scaled by the reciprocal of gear efficiency.
  \item \emph{Copper losses} ($b_i\Delta\tau_i^2$): resistive heating in the
        motor windings, proportional to the change in squared torque relative
        to the baseline.
  \item \emph{Coulomb friction} ($c_i|\dot{q}_i|$): constant friction
        dissipation, direction-independent and velocity-sign dependent in
        magnitude.
  \item \emph{Viscous friction} ($d_i\dot{q}_i^2$): velocity-dependent
        friction.
\end{itemize}
The $e_{ij}$ terms capture electromagnetic and mechanical coupling between
simultaneously moving joints, which is not accounted for by the
single-joint model alone. The identified parameter values for all joints are listed in
Table~\ref{tab:energy_params}. For more details, the reader is referred to \cite{DenizPowerModel2025}.

\begin{table}[t]
  \caption{Identified power model parameters for all 7 left-arm joints
           ($R^2 = 0.933$, $\mathrm{RMSE} = \SI{1.07}{\watt}$,
           hold-out $R^2 = 0.965$~\cite{DenizPowerModel2025}).
           Shoulder/elbow joints: copper loss ($b_i$) dominant.
           Wrist joints: viscous friction ($d_i$) dominant.}
  \label{tab:energy_params}
  \centering
  \setlength{\tabcolsep}{4pt}
  \begin{tabular}{lccccc}
    \toprule
    Joint & $i$ & $a_i$ & $b_i$ & $c_i$ & $d_i$ \\
    \midrule
    Shoulder pitch & 0 & 1.0 & 0.4184 & 0.0000 & 0.6467 \\
    Shoulder roll  & 1 & 1.0 & 0.3563 & 1.2767 & 0.0000 \\
    Shoulder yaw   & 2 & 1.0 & 0.2799 & 0.0000 & 0.2749 \\
    Elbow          & 3 & 1.0 & 0.3942 & 0.0324 & 0.0000 \\
    Wrist roll     & 4 & 1.0 & 0.0000 & 0.0000 & 1.0282 \\
    Wrist pitch    & 5 & 1.0 & 0.3768 & 0.3464 & 1.2004 \\
    Wrist yaw      & 6 & 1.0 & 0.0000 & 0.0000 & 1.7848 \\
    \bottomrule
  \end{tabular}

  \vspace{3pt}
  \footnotesize{Units: $b_i$ in \si{\watt\per\newton\squared\metre\squared},
  $c_i$ in \si{\watt\second\per\radian}, $d_i$ in \si{\watt\second\squared\per\radian\squared}.}

  \vspace{4pt}
  \begin{tabular}{lcc}
    \toprule
    Pair $(i,j)$ & $e_{ij}$ & Pair $(i,j)$ \\
    \midrule
    $(0,3)$ & 0.909 & $(1,6)$: 0.511 \\
    $(1,3)$ & 1.105 & $(2,6)$: 0.861 \\
    $(0,6)$ & 0.437 & $(2,3)$: 0.405 \\
    \multicolumn{3}{l}{\footnotesize{All other $e_{ij} < 0.12$ (see~\cite{DenizPowerModel2025}).}} \\
    \bottomrule
  \end{tabular}
\end{table}

% =============================================================================
\section{Reinforcement Learning Formulation}
\label{sec:rl}

\subsection{Markov Decision Process}
\label{sec:mdp}

We model the arm reaching task as a discrete-time Markov Decision Process
(MDP) $\mathcal{M} = (\mathcal{S}, \mathcal{A}, \mathcal{T}, r, \gamma)$,
with a fixed time step $\Delta t = \SI{0.01}{\second}$ and episode length
$T = 300$ steps (\SI{3.0}{\second}).

\subsubsection*{State Space}
The state $\vect{s}_t \in \R^{21}$ is
\begin{equation}
  \label{eq:state}
  \vect{s}_t =
  \Bigl[
    \tilde{\vect{q}}_t,\;
    \tilde{\dot{\vect{q}}}_t,\;
    \hat{\vect{e}}_p^t,\;
    \hat{\vect{e}}_o^t,\;
    t_{\mathrm{rem}}
  \Bigr],
\end{equation}
where:
\begin{itemize}
  \item $\tilde{\vect{q}}_t \in [-1,1]^7$ are joint angles normalised by their
        range: $\tilde{q}_i = 2(q_i - \bar{q}_i)/\Delta q_i$, with
        $\bar{q}_i = (q_i^{\max}+q_i^{\min})/2$ and
        $\Delta q_i = q_i^{\max}-q_i^{\min}$.
  \item $\tilde{\dot{\vect{q}}}_t = \dot{\vect{q}}_t / \dot{q}_{\max} \in \R^7$.
  \item $\vect{e}_p^t = \vect{p}^* - \vect{p}_t \in \R^3$ is the Cartesian
        position error of the end-effector (\si{\metre}), and
        $\hat{\vect{e}}_p^t = \vect{e}_p^t / d_{\mathrm{norm}}$ with
        $d_{\mathrm{norm}} = \SI{0.5}{\metre}$.
  \item $\vect{e}_o^t \in \so$ is the orientation error expressed as a
        rotation vector (axis $\times$ angle):
        \begin{equation}
          \vect{e}_o^t = \mathrm{Log}_{\SO}\!\left(\mat{R}^*
          \mat{R}_t^\top\right),
        \end{equation}
        where $\mat{R}^*$ and $\mat{R}_t \in \SO$ are the target and current
        end-effector rotation matrices, respectively, and
        $\hat{\vect{e}}_o^t = \vect{e}_o^t / \pi$.
  \item $t_{\mathrm{rem}} = 1 - k/T \in [0,1]$ is the fraction of time
        remaining in the episode.
\end{itemize}
End-effector pose $(\vect{p}_t, \mat{R}_t)$ is obtained from forward
kinematics.
% via the \texttt{left\_wrist\_yaw\_link} frame.

\subsubsection*{Action Space: Incremental Joint-Position Targets}
\label{sec:action_space}
% An earlier formulation in which the action directly specified a normalised
% joint-velocity command produced policies whose commands saturated on
% essentially every step (Section~\ref{sec:lineage}).  
We use an \emph{incremental joint-position} action space, in which the action
$\vect{a}_t \in [-1,1]^7$ specifies a displacement of a position target that is
then tracked by a proportional joint controller:
\begin{align}
  \vect{q}_t^{\mathrm{tgt}}
    &= \mathrm{clip}\!\left(\vect{q}_t + \vect{a}_t \odot \Delta q_{\max},\;
       \vect{q}^{\min},\,\vect{q}^{\max}\right),
       \label{eq:q_target}\\
  \dot{\vect{q}}_t^{\mathrm{des}}
    &= k_p\left(\vect{q}_t^{\mathrm{tgt}} - \vect{q}_t\right)
       - k_d\,\dot{\vect{q}}_t,
       \label{eq:dq_desired}\\
  \dot{\vect{q}}_t^{\mathrm{cmd}}
    &= \mathrm{clip}\!\left(\dot{\vect{q}}_t^{\mathrm{des}},\;
       -\dot{q}_{\max},\,\dot{q}_{\max}\right),
       \label{eq:dq_cmd}
\end{align}
where $\Delta q_{\max} = \dot{q}_{\max}/k_p \approx \SI{0.1667}{\radian}$ is
the maximum per-step position increment, $k_p = 15.0$ is the position gain,
and $k_d = 0$.  Bounding the position increment by $\dot{q}_{\max}/k_p$
guarantees that the desired velocity~\eqref{eq:dq_desired} cannot exceed the
joint speed limit even before clipping in~\eqref{eq:dq_cmd}, which removes the
persistent saturation observed with direct velocity actions.

\subsubsection*{Transition Dynamics}
Joint positions are integrated with the Euler scheme subject to joint limits:
\begin{equation}
  \label{eq:q_next}
  \vect{q}_{t+1} = \mathrm{clip}\!\left(
  \vect{q}_t + \dot{\vect{q}}_t^{\mathrm{cmd}}\,\Delta t,\;
  \vect{q}^{\min},\; \vect{q}^{\max}\right),
\end{equation}
and the realised velocity is $\dot{\vect{q}}_{t+1} = (\vect{q}_{t+1} -
\vect{q}_t)/\Delta t$.  Joint torques are computed via the Recursive
Newton-Euler Algorithm (RNEA)~\cite{Featherstone2008} as implemented in
Pinocchio~\cite{Carpentier2019}, evaluated at the \emph{pre}-update state
$(\vect{q}_t,\dot{\vect{q}}_t)$ with the realised acceleration:
\begin{equation}
  \label{eq:rnea}
  \vect{\tau}_t =
  \mathrm{RNEA}\!\left(\vect{q}_t,\,\dot{\vect{q}}_t,\,
  \ddot{\vect{q}}_t\right),
  \quad
  \ddot{\vect{q}}_t =
  \frac{\dot{\vect{q}}_{t+1} - \dot{\vect{q}}_t}{\Delta t}.
\end{equation}
A reduced 7-DOF Pinocchio model is built from the full 29-DOF URDF by locking
all non-arm joints at their neutral configuration, so RNEA runs at the arm
level without the computational cost of the full-body model.

\subsection{Hybrid Constellation Reward}
\label{sec:reward}

% A purely Cartesian reward that penalises $\norm{\vect{e}_p^t}$ and
% $\norm{\vect{e}_o^t}$ separately, as used in earlier versions of this work,
% requires balancing two terms with very different units and gradients, and was
% found empirically to under-weight orientation alignment relative to position.
We define a \emph{constellation distance} that combines position and
orientation error into a single geometric quantity by tracking a small set of
virtual points rigidly attached to the end-effector frame.

\subsubsection*{Constellation Distance}
Let $N_c = 4$ constellation points be placed at angles
$\theta_k = 2\pi k / N_c$, $k=0,\dots,N_c-1$, on a circle of radius
$r_c = \SI{0.12}{\metre}$ in the end-effector's local $xy$-plane:
\begin{equation}
  \label{eq:constellation_points}
  \vect{c}_k = r_c \left[\cos\theta_k,\ \sin\theta_k,\ 0\right]^\top.
\end{equation}
At step $t$, the current and target positions of constellation point $k$ are
\begin{align}
  \vect{x}_k^t &= \vect{p}_t + \mat{R}_t\,\vect{c}_k, \label{eq:con_current}\\
  \vect{x}_k^* &= \vect{p}^* + \mat{R}^*\,\vect{c}_k, \label{eq:con_target}
\end{align}
and the constellation distance is their mean squared separation,
\begin{equation}
  \label{eq:d_con}
  d_{\mathrm{con}}^t = \frac{1}{N_c}\sum_{k=0}^{N_c-1}
  \norm{\vect{x}_k^t - \vect{x}_k^*}^2 .
\end{equation}
Because the constellation points are offset from the end-effector origin,
$d_{\mathrm{con}}^t$ is sensitive to \emph{both} the translational error
$\vect{e}_p^t$ and the rotational misalignment between $\mat{R}_t$ and
$\mat{R}^*$: a pure rotation about $\vect{p}_t$ with $\vect{p}_t = \vect{p}^*$
still displaces the constellation points and increases $d_{\mathrm{con}}^t$.
A single exponential or quadratic term in $d_{\mathrm{con}}^t$ therefore
shapes position \emph{and} orientation jointly, without separately tuned
weights for each.

\subsubsection*{Reward}
The reward at step $t$ is the sum of six terms,
\begin{align}
  r_t =\;
  &\delta_{\mathrm{con}}\left(d_{\mathrm{con}}^{t-1} - d_{\mathrm{con}}^{t}\right)
  \label{eq:r_progress}\\
  &- \beta_{\mathrm{res}}\,\norm{\vect{e}_p^t}
  \label{eq:r_res}\\
  &+ \exp\!\left(-w_{\mathrm{con}}\,d_{\mathrm{con}}^t\right)
  \label{eq:r_con}\\
  &- c_{\mathrm{step}}
  \label{eq:r_step}\\
  &- \lambda_{\mathrm{smooth}}\,\norm{\dot{\vect{q}}_t - \dot{\vect{q}}_{t-1}}^2
  \label{eq:r_smooth}\\
  &- \lambda_\tau\,\norm{\vect{\tau}_t}^2
  \label{eq:r_tau}\\
  &+ R_{\mathrm{success}}\,\mathbf{1}[\text{success}],
  \label{eq:r_success}
\end{align}
where
$\text{success} \triangleq \left(\norm{\vect{e}_p^t} < d_p\right)
\wedge \left(\norm{\vect{e}_o^t} < d_o\right)$ terminates the episode.  Term by
term:
\begin{itemize}
  \item \eqref{eq:r_progress}: \emph{progress shaping}---a dense reward
        proportional to the reduction in constellation distance since the
        previous step, weight $\delta_{\mathrm{con}} = 5.0$.
  \item \eqref{eq:r_res}: \emph{residual position penalty}---a small linear
        penalty on the remaining Cartesian distance, weight
        $\beta_{\mathrm{res}} = 0.1$, which breaks ties between trajectories
        with similar progress but different absolute distance to the target.
  \item \eqref{eq:r_con}: \emph{exponential constellation reward}---a
        bounded, strictly positive term in $(0,1]$ that grows sharply as
        $d_{\mathrm{con}}^t \to 0$, with sharpness $w_{\mathrm{con}} = 40.0$.
        Unlike~\eqref{eq:r_progress}, this term rewards \emph{proximity}
        regardless of the path taken to get there, preventing the agent from
        exploiting the progress term through oscillatory motion.
  \item \eqref{eq:r_step}: a constant per-step cost $c_{\mathrm{step}} = 0.001$
        that encourages the agent to terminate (reach the goal) as early as
        possible.
  \item \eqref{eq:r_smooth}: a \emph{velocity-smoothness} penalty (active for
        $t>0$), weight $\lambda_{\mathrm{smooth}} = 0.005$, discouraging
        abrupt changes in realised velocity between consecutive steps.
  \item \eqref{eq:r_tau}: a \emph{torque-norm energy proxy}, weight
        $\lambda_\tau = 5\times10^{-5}$ (see Remark~\ref{rem:energy_proxy}).
  \item \eqref{eq:r_success}: a sparse terminal bonus
        $R_{\mathrm{success}} = 700.0$, awarded when the end-effector is
        simultaneously within $d_p = \SI{4}{\centi\metre}$ position and
        $d_o = 0.15\,\mathrm{rad} \approx 8.6^\circ$ orientation tolerance of
        the target.
\end{itemize}

\begin{remark}[Energy term: reward proxy vs.\ evaluation metric]
\label{rem:energy_proxy}
The full electrical power model $P_{\mathrm{net}}(\vect{\tau}_t,
\dot{\vect{q}}_t)$~\eqref{eq:power} is evaluated and logged at every training
step, but it does \emph{not} appear in the reward: the corresponding weight
$\alpha$ is set to $0$.  The only energy-related term that shapes the policy
during training is the torque-norm proxy~\eqref{eq:r_tau},
$\lambda_\tau\norm{\vect{\tau}_t}^2$, which is cheap to compute, smooth in
$\vect{\tau}_t$, and penalises the large inertial torques produced by abrupt
accelerations without requiring the full nonlinear power model inside the
training loop.  The physically grounded power model~\eqref{eq:power} is
instead applied \emph{post hoc} to the realised $(\vect{\tau}_t,\dot{\vect{q}}_t)$
trajectory of every evaluation episode---in kinematic simulation, in MuJoCo,
and on the physical robot---and is the metric reported throughout
Section~\ref{sec:experiments}.  This separation keeps the reward landscape
smooth and inexpensive during the training run while
preserving the experimentally calibrated power model as the common basis for
all energy comparisons.
\end{remark}

\subsection{Soft Actor-Critic}
\label{sec:sac}

We adopt Soft Actor-Critic (SAC)~\cite{Haarnoja2018}, an off-policy
maximum-entropy RL algorithm well-suited to continuous action spaces and
shaped reward landscapes.  SAC optimises the entropy-augmented objective
\begin{equation}
  \label{eq:sac_objective}
  J(\pi) = \mathbb{E}_{\tau\sim\pi}
  \left[\sum_{t=0}^{T} \gamma^t
  \Bigl( r_t + \alpha_{\mathrm{ent}}\,\mathcal{H}(\pi(\cdot|\vect{s}_t)) \Bigr)
  \right],
\end{equation}
where $\mathcal{H}$ is the policy entropy and $\alpha_{\mathrm{ent}}$ is an
automatically tuned temperature parameter~\cite{Haarnoja2018auto}.  The
entropy bonus plays a critical role here: because the success region defined
by $(d_p, d_o)$ in~\eqref{eq:r_success} is a small volume in joint space, the
agent must maintain broad exploration even near the goal to discover terminal
states that satisfy both the position and orientation tolerances
simultaneously, rather than committing prematurely to a suboptimal
approach direction.

The policy and both critics are multi-layer perceptrons (MLPs) with
architecture $[256, 256]$.  Algorithm~\ref{alg:sac} summarises one
environment step of the training loop, combining the incremental
action-to-torque mapping of Section~\ref{sec:action_space} with the standard
SAC update.  Table~\ref{tab:sac_hyperparams} lists all hyperparameters.

\begin{algorithm}[t]
\caption{SAC training step with incremental position-target action and
RNEA-based torque evaluation}
\label{alg:sac}
\begin{algorithmic}[1]
\STATE \textbf{Input:} policy $\pi_\theta$; critics $Q_{\phi_1},Q_{\phi_2}$
       and targets $Q_{\phi_1'},Q_{\phi_2'}$; replay buffer $\mathcal{D}$
\FOR{environment step $t = 0,1,\dots$}
  \STATE Observe state $\vect{s}_t$~\eqref{eq:state}
  \STATE Sample action $\vect{a}_t \sim \pi_\theta(\cdot|\vect{s}_t)$
  \STATE Compute position target $\vect{q}_t^{\mathrm{tgt}}$~\eqref{eq:q_target}
  \STATE Compute commanded velocity $\dot{\vect{q}}_t^{\mathrm{cmd}}$
         via~\eqref{eq:dq_desired}--\eqref{eq:dq_cmd}
  \STATE Integrate $\vect{q}_{t+1}$~\eqref{eq:q_next};
         $\dot{\vect{q}}_{t+1} \leftarrow (\vect{q}_{t+1}-\vect{q}_t)/\Delta t$
  \STATE Evaluate $\vect{\tau}_t \leftarrow
         \mathrm{RNEA}(\vect{q}_t,\dot{\vect{q}}_t,\ddot{\vect{q}}_t)$~\eqref{eq:rnea}
  \STATE Compute $d_{\mathrm{con}}^t$~\eqref{eq:d_con} and reward $r_t$
         via~\eqref{eq:r_progress}--\eqref{eq:r_success}
  \STATE Log $P_{\mathrm{net}}(\vect{\tau}_t,\dot{\vect{q}}_t)\,\Delta t$
         \eqref{eq:power} (evaluation only, not part of $r_t$)
  \STATE Store $(\vect{s}_t,\vect{a}_t,r_t,\vect{s}_{t+1},\text{done})$ in $\mathcal{D}$
  \IF{$t \geq$ \texttt{learning\_starts}}
    \FOR{$g = 1$ \TO \texttt{gradient\_steps}}
      \STATE Sample minibatch from $\mathcal{D}$
      \STATE Update $Q_{\phi_1}, Q_{\phi_2}$ by minimising the soft Bellman
             residual using $Q_{\phi_1'}, Q_{\phi_2'}$
      \STATE Update $\pi_\theta$ by maximising
             $\mathbb{E}\!\left[\min_i Q_{\phi_i}(\vect{s},\vect{a}) -
             \alpha_{\mathrm{ent}}\log\pi_\theta(\vect{a}|\vect{s})\right]$
      \STATE Update $\alpha_{\mathrm{ent}}$ toward the target entropy
      \STATE Soft-update targets:
             $\phi_i' \leftarrow \tau_{\mathrm{soft}}\phi_i +
             (1-\tau_{\mathrm{soft}})\phi_i'$
    \ENDFOR
  \ENDIF
\ENDFOR
\end{algorithmic}
\end{algorithm}

\begin{table}[t]
  \caption{SAC hyperparameters}
  \label{tab:sac_hyperparams}
  \centering
  \begin{tabular}{lc}
    \toprule
    Hyperparameter & Value \\
    \midrule
    Learning rate              & $3\times10^{-4}$ \\
    Discount factor $\gamma$   & 0.99 \\
    Soft update $\tau_{\mathrm{soft}}$ & 0.005 \\
    Replay buffer size         & $5\times10^5$ \\
    Batch size                 & 256 \\
    Learning starts            & 25\,000 \\
    Train frequency            & 1 step \\
    Gradient steps per update  & 4 \\
    Temperature $\alpha_{\mathrm{ent}}$ & auto \\
    Network architecture (policy \& critics) & $[256, 256]$ \\
    \bottomrule
  \end{tabular}
\end{table}

% =============================================================================
\section{Training Setup}
\label{sec:training}

\subsection{Simulation Environment}

Training runs entirely in the kinematic Pinocchio environment of
Section~\ref{sec:mdp}.  Environments are instantiated with the
Stable-Baselines3~\cite{Raffin2021} \texttt{SubprocVecEnv} wrapper using
$N_{\mathrm{env}} = 8$ parallel processes, yielding an effective wall-clock
throughput of approximately $1\,000$--$1\,700$ steps per second on a desktop
GPU (NVIDIA RTX series).
A \texttt{VecNormalize} layer is intentionally omitted because the observation
terms are already normalised to approximately $[-1, 1]$ by construction
(Section~\ref{sec:mdp}).

\subsection{Target Sampling and Initial State}
\label{sec:target_sampling}

At each episode reset, a target pose $(\vect{p}^*, \mat{R}^*)$ is sampled by
drawing a random joint configuration $\vect{q}^* \sim
\mathcal{U}(\vect{q}^{\min},\vect{q}^{\max})$ and computing its forward
kinematics; the sample is rejected and redrawn if $\norm{\vect{p}^*} \leq
\SI{0.15}{\metre}$, which excludes degenerate target poses close to the
shoulder.  No further bias toward any workspace region is applied: $\vect{q}^*$
is accepted as soon as the distance check passes.

The initial joint configuration is drawn from a band around the neutral pose,
\begin{equation}
  \vect{q}_0 \sim \mathcal{U}\!\left(
    \mathrm{clip}(\vect{q}_c - 0.15\,\Delta\vect{q},\,\vect{q}^{\min}),\;
    \mathrm{clip}(\vect{q}_c + 0.15\,\Delta\vect{q},\,\vect{q}^{\max})
  \right),
\end{equation}
with $\vect{q}_c = (\vect{q}^{\max}+\vect{q}^{\min})/2$ and
$\Delta\vect{q} = \vect{q}^{\max}-\vect{q}^{\min}$, and
$\dot{\vect{q}}_0 = \vect{0}$.

\subsection{Training Lineage and Protocol}
\label{sec:lineage}

The agent is trained for a total of $5\times10^6$ environment steps.  An
\texttt{EvalCallback} evaluates the deterministic policy on held-out episodes
periodically and saves the checkpoint achieving the highest mean reward; a
\texttt{CheckpointCallback} additionally saves snapshots every $10^5$ steps.
Each fresh run discards the replay buffer to avoid distributional shift when
the MDP or reward changes.

\subsection{Baseline: Minimum-Jerk Trajectory}
\label{sec:minjerk_baseline}

As a reference for simulation evaluation, we use a joint-space minimum-jerk
trajectory~\cite{Flash1985}: a 5th-order polynomial applied directly between
the initial joint configuration and the sampled target configuration
$\vect{q}^*$ (see Appendix~\ref{app:minjerk}).  No inverse kinematics is
needed because the target is defined in joint space via forward kinematics
sampling, so the same target is available to both the RL policy (as a
Cartesian pose) and the min-jerk planner (as a joint configuration).  The
trajectory duration is set to satisfy the velocity limit
$\dot{q}_{\max} = \SI{2.5}{\radian\per\second}$ for the largest joint
displacement, giving a planned duration of $1.519\pm0.342$\,s
(Appendix~\ref{app:minjerk})---well within the
\SI{3.0}{\second} RL episode budget.  The realised episode duration
(Table~\ref{tab:sim_results}) is shorter on average ($1.276\pm0.357$\,s)
because episodes terminate as soon as the success tolerance is reached,
before the planned trajectory completes.  The min-jerk baseline is energy-unaware;
energy consumption is computed after the fact by passing the resulting
trajectory through the same RNEA and identified power model used to evaluate
the RL policy.

\subsection{MuJoCo Dynamics Validation}
\label{sec:mujoco_validation}

The kinematic environment of Section~\ref{sec:mdp} assumes a
\emph{perfect tracker}: the commanded velocity $\dot{\vect{q}}_t^{\mathrm{cmd}}$
is realised exactly within one step~\eqref{eq:q_next}.  Before any hardware
trial, we therefore validate the frozen $69.9\%$ policy (no further training)
in MuJoCo~\cite{Todorov2012} using the full Unitree~G1 MJCF model (29~DOF,
floating base).  At every $\Delta t = \SI{10}{\milli\second}$ RL step, the
simulator advances 5 physics substeps of \SI{2}{\milli\second}; the floating
base and the 36 non-arm joints are teleport-locked to their default
configuration $\vect{q}_0^{\mathrm{mj}}$ after every substep, isolating the
7-DOF left arm under realistic actuator dynamics without requiring a balance
controller.  At each RL step the policy's position
target~\eqref{eq:q_target} and commanded velocity~\eqref{eq:dq_cmd} are passed
to a low-level PD torque controller,
\begin{equation}
  \label{eq:pd_torque}
  \vect{\tau}_t^{\mathrm{cmd}} = k_p^{\mathrm{mj}}\left(\vect{q}_t^{\mathrm{tgt}}
  - \vect{q}_t^{\mathrm{mj}}\right)
  + k_d^{\mathrm{mj}}\left(\dot{\vect{q}}_t^{\mathrm{cmd}} -
  \dot{\vect{q}}_t^{\mathrm{mj}}\right),
\end{equation}
clipped to the MJCF actuator \texttt{ctrlrange}
($\pm\SI{25}{\newton\metre}$ for shoulder/elbow/wrist-roll,
$\pm\SI{5}{\newton\metre}$ for wrist-pitch/yaw)---tighter than the
RNEA-evaluation torque limits used in Section~\ref{sec:rl}.

\subsubsection*{PD-Gain Sweep}
\begin{figure}[t]
  \centering
  \includegraphics[width=\columnwidth]{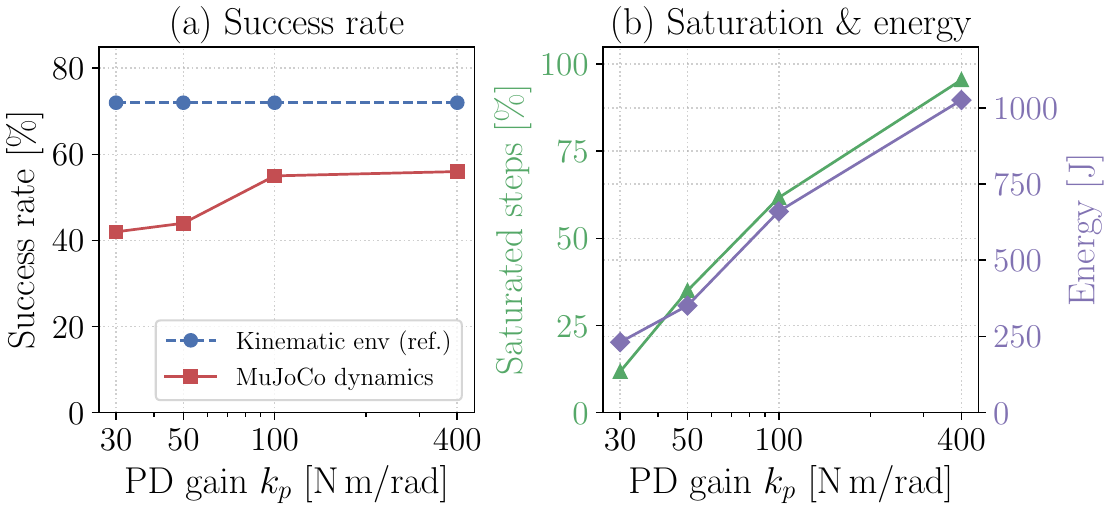}
  \caption{Frozen-policy MuJoCo validation across the PD-gain sweep.
           (a)~Success rate vs.\ $k_p^{\mathrm{mj}}$: the kinematic-environment
           rate (dashed) is gain-independent by construction, while the
           MuJoCo dynamics rate is consistently $20$--$25$ percentage points
           lower.  (b)~Actuator saturation and mean episode energy both grow
           sharply with $k_p^{\mathrm{mj}}$.}
  \label{fig:kp_sweep}
\end{figure}

Because $k_p^{\mathrm{mj}}$ and $k_d^{\mathrm{mj}}$ are deployment choices not
seen during training, we sweep $k_p^{\mathrm{mj}} \in \{30, 50, 100, 400\}$
(with $k_d^{\mathrm{mj}} = k_p^{\mathrm{mj}}/20$) over $n=100$ targets and
report the kinematic-environment success rate on the same targets as a
reference.  Table~\ref{tab:kp_sweep} and Fig.~\ref{fig:kp_sweep} summarise the
results; an additional $n=200$ run at $k_p^{\mathrm{mj}}=400$ confirms the
trend at larger sample size.

\begin{table}[t]
  \caption{MuJoCo frozen-policy PD-gain sweep ($n=100$ unless noted).
           ``Sat.''\ is the fraction of action-clipped steps; ``ori.\ err.''\
           is the mean final orientation error.}
  \label{tab:kp_sweep}
  \centering
  \begin{tabular}{lccccc}
    \toprule
    $k_p^{\mathrm{mj}}$ & Kin.\ succ.\ (\%) & Dyn.\ succ.\ (\%) & Sat.\ (\%) &
    Energy (J) & Ori.\ err.\ ($^\circ$) \\
    \midrule
    30  & 72 & 42 & 11.8 & 231  & 11.0 \\
    50  & 72 & 44 & 35.0 & 351  & 11.2 \\
    100 & 72 & 55 & 61.7 & 660  & 11.8 \\
    400 & 72 & 56 & 95.5 & 1025 & 13.4 \\
    \midrule
    \multicolumn{6}{l}{\footnotesize $400$ ($n=200$): Kin.\ $71$, Dyn.\ $46$, Sat.\ $94.7$, Energy $1110$, Ori.\ $13.4$}\\
    \bottomrule
  \end{tabular}
\end{table}

Two findings drive the deployment choices used in
Section~\ref{sec:experiments}.  First, the success-rate gap between the
kinematic environment and MuJoCo is $20$--$25$ percentage points across
\emph{all} tested gains---i.e., it is largely \emph{gain-independent}---and
the mean final orientation error is consistently $11$--$13^\circ$, roughly
$3$--$5^\circ$ above the $8.6^\circ$ training tolerance, regardless of
$k_p^{\mathrm{mj}}$.  This indicates that the gap is not primarily a
controller-tuning problem but an intrinsic consequence of the
perfect-tracker assumption in Section~\ref{sec:mdp}: the policy was trained
assuming instantaneous, exact velocity tracking, and any finite-gain PD
controller introduces a residual tracking lag that the policy never
experienced.  Second, actuator saturation and energy consumption are highly
sensitive to $k_p^{\mathrm{mj}}$, growing roughly $8\times$ from
$k_p^{\mathrm{mj}}=30$ to $400$.  We therefore deploy with a soft PD
configuration ($k_p^{\mathrm{mj}}=50$, $k_d^{\mathrm{mj}}=1.5$) together with
the torque safety clamp above, which keeps saturation and energy near the
lower end of the sweep without sacrificing the success rate achievable at
higher gains.  The remaining $\sim20$-point gap is addressed in
Section~\ref{sec:workspace} through a workspace-reachability analysis that
identifies \emph{where} in the workspace this gap is concentrated.

% =============================================================================
\section{Experimental Results}
\label{sec:experiments}

\begin{figure}[t]
  \centering
  \includegraphics[width=\columnwidth]{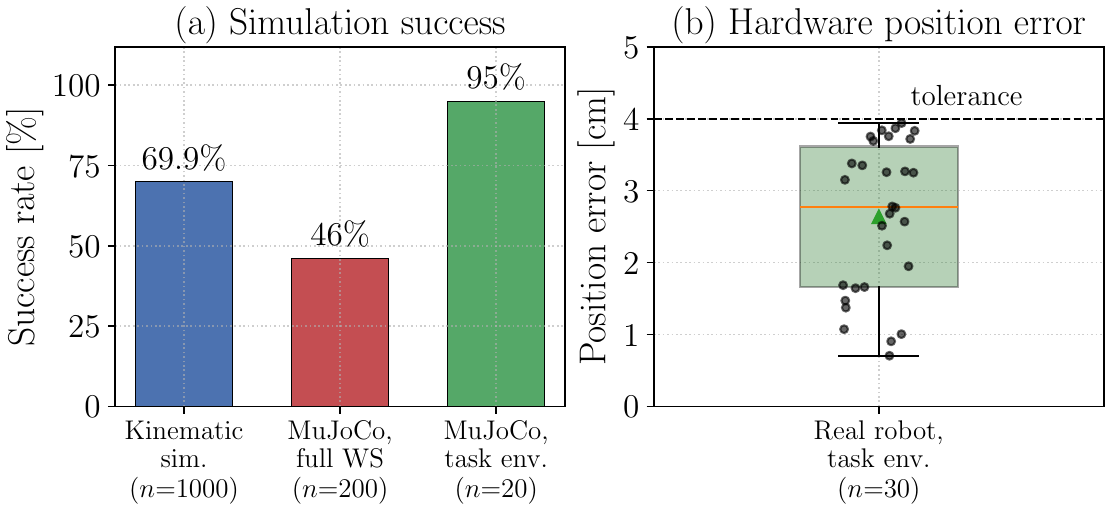}
  \caption{Validation arc summary.  (a)~Success rate in simulation: the
           kinematic-simulation rate ($69.9\%$, $n{=}1000$) drops to $46\%$
           under MuJoCo dynamics over the full nominal workspace ($n{=}200$,
           Section~\ref{sec:mujoco_validation}), and recovers to $95\%$
           ($n{=}20$) once evaluation is restricted to the reachable
           apple-picking task envelope (Section~\ref{sec:workspace}).
           (b)~On the physical Unitree~G1 ($n{=}30$, same task envelope), we
           report the achieved end-effector position error directly: every
           trial lands below the \SI{4}{\centi\metre} tolerance (dashed
           line), with mean $2.64\pm1.04$\,cm and worst case $3.94$\,cm.}
  \label{fig:success_funnel}
\end{figure}

The RL formulation of Section~\ref{sec:rl} and the training/validation
pipeline of Section~\ref{sec:training} are evaluated along a four-stage arc,
summarised in Fig.~\ref{fig:success_funnel}: (i)~kinematic-simulation
evaluation of the frozen $69.9\%$ policy at $n=1\,000$
(Section~\ref{sec:sim_results}); (ii)~a workspace-reachability analysis that
identifies a restricted task envelope in which the policy can be expected to
transfer, together with an $n=20$ MuJoCo screening on that envelope
(Section~\ref{sec:workspace}); and (iii)~real-hardware validation on the
physical Unitree~G1 over $n=30$ independent trials
(Section~\ref{sec:hardware}).  The MuJoCo dynamics-validation methodology and
the PD-gain sweep that motivate stages (ii)--(iii) were already presented in
Section~\ref{sec:mujoco_validation}.  For the first three, simulation-based
stages, a binary success rate over a fixed tolerance is a meaningful summary
statistic because $n$ is large ($1\,000$, $200$, and $20$, respectively); for
the real-robot stage, $n=30$ is too small for a binary rate to be statistically
informative, so we instead report the full distribution of end-effector
position errors achieved.

\subsection{Kinematic Simulation Evaluation (\texorpdfstring{$n=1\,000$}{n=1,000})}
\label{sec:sim_results}

The frozen reference policy (Section~\ref{sec:lineage})
is evaluated deterministically over $n=1\,000$ targets sampled as in
Section~\ref{sec:target_sampling}, against the minimum-jerk baseline of
Section~\ref{sec:minjerk_baseline} on the same targets.
Table~\ref{tab:sim_results} reports overall metrics; Table~\ref{tab:sim_perjoint}
reports the per-joint kinematic/torque statistics and the breakdown of the
post-hoc energy model~\eqref{eq:power} for the RL policy.

\begin{table}[t]
  \caption{Kinematic-simulation evaluation, $n=1\,000$ targets, deterministic
           policy (checkpoint at $5\times10^6$ steps).  Tolerances:
           $d_p=\SI{4}{\centi\metre}$, $d_o=8.6^\circ$.}
  \label{tab:sim_results}
  \centering
  \begin{tabular}{lcc}
    \toprule
    Metric & RL (SAC) & Min-jerk \\
    \midrule
    Success rate (\%)                       & $69.9$              & $100.0$ \\
    Energy, all episodes (J)                & $163.7 \pm 838.3$   & $22.0 \pm 9.6$ \\
    Energy, successes only (J)              & $98.16$ ($n{=}699$) & $21.95$ ($n{=}1000$) \\
    Episode duration (s)                    & $1.542 \pm 1.049$   & $1.276 \pm 0.357$ \\
    Final pos.\ error, all (cm)             & $6.08 \pm 7.32$     & $2.50 \pm 0.96$ \\
    Final ori.\ error, all ($^\circ$)       & $16.49 \pm 24.47$   & $7.88 \pm 0.93$ \\
    Smoothness MSJ (rad$^2$ s$^{-6}$)        & $1.31\times10^{7}$  & $2.00\times10^{2}$ \\
    Action saturation (\% steps)            & $24.6$              & $12.1$ \\
    \bottomrule
  \end{tabular}
\end{table}

\begin{table*}[t]
  \caption{RL policy, $n=1\,000$: per-joint kinematics/torques (mean over all
           episodes) and post-hoc energy breakdown~\eqref{eq:power} (mean
           over all episodes).}
  \label{tab:sim_perjoint}
  \centering
  \begin{tabular}{lccccc}
    \toprule
    Joint & $|\dot q|$ mean (rad/s) & $|\dot q|$ max & Saturation (\%) &
    $|\tau|$ mean (N\,m) & $|\tau|$ max (N\,m) \\
    \midrule
    Sh.\ pitch & 1.17 & 2.50 & 7.7 & 4.62  & 199.46 \\
    Sh.\ roll  & 0.93 & 2.50 & 4.1 & 4.23  & 159.31 \\
    Sh.\ yaw   & 1.10 & 2.50 & 5.5 & 2.04  &  64.24 \\
    Elbow      & 0.94 & 2.50 & 2.6 & 2.62  &  99.94 \\
    Wr.\ roll  & 1.09 & 2.50 & 5.1 & 0.59  &  27.75 \\
    Wr.\ pitch & 1.01 & 2.50 & 2.7 & 0.85  &  43.37 \\
    Wr.\ yaw   & 1.03 & 2.50 & 3.9 & 0.66  &  22.10 \\
    \bottomrule
  \end{tabular}
  \vspace{4pt}

  \begin{tabular}{lccccc}
    \toprule
    Energy term & Mechanical & Copper & Coulomb & Viscous & Interaction \\
    \midrule
    Mean (J) & 11.66 & 143.46 & 2.42 & 14.60 & 11.05 \\
    Share (\%) & 7.1 & 87.6 & 1.5 & 8.9 & 6.8 \\
    \bottomrule
  \end{tabular}
\end{table*}
\begin{remark}The minimum-jerk baseline succeeds
on $100\%$ of episodes (Table~\ref{tab:sim_results}), with final errors
clustered just inside the tolerance band ($2.50\pm0.96$\,cm,
$7.88\pm0.93^\circ$ vs.\ $d_p=\SI{4}{\centi\metre}$/$d_o=8.6^\circ$).
Because the closed-form trajectory~\eqref{eq:minjerk} approaches
$\vect{q}^*$ monotonically with vanishing velocity and acceleration at
$\xi=1$, the success criterion is met at $\xi<1$ on average, so episodes
terminate early (mean $1.276\pm0.357$\,s vs.\ the planned
$1.519\pm0.342$\,s, Appendix~\ref{app:minjerk}).  Note that the planner is
constructed with full knowledge of $\vect{q}^*$---effectively a privileged
inverse-kinematics solution unavailable to the RL policy, which observes
only the Cartesian target pose---so Table~\ref{tab:sim_results} should be
read as a smooth, energy-efficient \emph{oracle} reference rather than a
directly comparable control strategy.
\end{remark}

On the RL policy, the energy breakdown in Table~\ref{tab:sim_perjoint} mirrors
the pattern of the power model itself (Table~\ref{tab:energy_params}): copper
losses dominate ($87.6\%$), consistent with the large peak torques (up to
$\SI{199}{\newton\metre}$ on the shoulder pitch joint) generated by
RNEA~\eqref{eq:rnea} during the fast, large-amplitude motions that the
kinematic environment's perfect-tracker assumption permits.  Action
saturation occurs on $24.6\%$ of steps, concentrated on the shoulder joints
(up to $7.7\%$ per joint), reflecting the incremental action space's
displacement ceiling~\eqref{eq:q_target} being reached during the most
aggressive phases of a reach.

\subsection{Workspace Reachability and Restricted Task Envelope}
\label{sec:workspace}

Section~\ref{sec:mujoco_validation} showed that the kinematic-to-MuJoCo
success gap is approximately $20$--$25$ percentage points and largely
gain-independent.  To determine how much of this gap is explained by targets
that are simply not reachable by the arm---independent of any controller---we
sample $n=80$ random points inside the nominal apple-picking workspace
bounding box and solve unregularised position inverse kinematics (IK,
$30$ random restarts) for each.  Table~\ref{tab:reachability} (top) summarises
the result: only $24/80$ ($30\%$) of points are reachable within
$\SI{2}{\centi\metre}$; the remaining points miss by a median of
$\sim\SI{10}{\centi\metre}$ (90th percentile $\sim\SI{31}{\centi\metre}$, max
$\sim\SI{41}{\centi\metre}$), regardless of the IK initialisation or
controller---no feedback law can close a $10$+\,cm kinematic gap.  A second,
qualitatively distinct failure mode occurs for targets close to the torso
($x<\SI{0.1}{\metre}$ in the robot's forward direction): the arm can become
physically stuck in a history-dependent configuration up to $\sim 54^\circ$
from the commanded target, a mode not present in the kinematic training
environment.

\begin{table}[t]
  \caption{Workspace reachability ($n=80$ random points in the apple-picking
           bounding box, position IK with 30 restarts) and MuJoCo screening
           on the restricted reachable task envelope ($x \geq
           \SI{0.1}{\metre}$, IK residual $<\SI{2}{\centi\metre}$, $n=20$,
           default policy rollout, $k_p^{\mathrm{mj}}=50$,
           $k_d^{\mathrm{mj}}=1.5$).}
  \label{tab:reachability}
  \centering
  \begin{tabular}{lc}
    \toprule
    \multicolumn{2}{l}{\emph{Workspace reachability ($n=80$)}} \\
    \midrule
    Reachable within \SI{2}{\centi\metre}     & $24/80$ ($30\%$) \\
    Median miss (unreachable points)          & $\sim\SI{10}{\centi\metre}$ \\
    90th percentile miss                      & $\sim\SI{31}{\centi\metre}$ \\
    Maximum miss                              & $\sim\SI{41}{\centi\metre}$ \\
    \midrule
    \multicolumn{2}{l}{\emph{Restricted-envelope MuJoCo screening ($n=20$)}} \\
    \midrule
    Success rate                              & $19/20$ ($95\%$) \\
    Pos.\ error, median (range)               & $3.0$ ($1.3$--$6.1$)\,cm \\
    Ori.\ error, median (range)               & $7.0$ ($2.3$--$11.8$)$^\circ$ \\
    Mean motor excitation, range              & $0.05$--$0.26$ \\
    \bottomrule
  \end{tabular}
\end{table}

Restricting evaluation to the subset of the apple-picking workspace that is
(a)~at least \SI{0.1}{\metre} from the torso along $x$ and (b)~IK-reachable
within \SI{2}{\centi\metre}---i.e., the \emph{task envelope} that a real
apple-picking deployment would target---and re-running the frozen policy in
MuJoCo with the deployment PD gains ($k_p^{\mathrm{mj}}=50$,
$k_d^{\mathrm{mj}}=1.5$) over $n=20$ targets recovers a $95\%$ success rate
(Table~\ref{tab:reachability}, bottom), with median final errors of
$\SI{3.0}{\centi\metre}$/$7.0^\circ$, both within tolerance, and mean motor
excitation in $[0.05, 0.26]$---a substantially tighter, lower-effort regime
than the full-workspace sweep of Table~\ref{tab:kp_sweep}.  The single failure
($x=\SI{0.105}{\metre}$, near the lower edge of the envelope) timed out at
$6.1$\,cm/$11.8^\circ$, just outside tolerance.  This restricted task envelope
is the target distribution used for the real-hardware validation in
Section~\ref{sec:hardware}.

\subsection{Real-Robot Hardware Validation (\texorpdfstring{$n=30$}{n=30})}
\label{sec:hardware}

\subsubsection{Deployment}

The frozen policy is deployed on the physical Unitree~G1 via a Python node
that reads joint state from a ROS2 topic at \SI{100}{\hertz}, assembles the
21-dimensional observation $\vect{s}_t$~\eqref{eq:state}, passes it through
the frozen network (forward pass $<\SI{1}{\milli\second}$ on CPU), and
publishes the resulting position target and commanded
velocity~\eqref{eq:q_target}--\eqref{eq:dq_cmd} to the low-level joint
controller via a ROS2 topic, using the deployment PD gains identified in
Section~\ref{sec:mujoco_validation} ($k_p^{\mathrm{mj}}=50$,
$k_d^{\mathrm{mj}}=1.5$) and the same torque safety clamp.  No additional
sim-to-real adaptation (domain randomisation, friction identification, or
fine-tuning) is applied.

\subsubsection{Protocol}

\begin{figure}[t]
  \centering
  \includegraphics[width=\columnwidth]{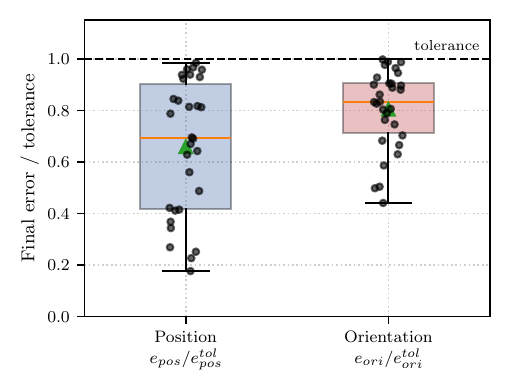}
  \caption{Final end-effector position and orientation error on hardware
           ($n=30$), normalised by the success tolerance
           ($d_p=\SI{4}{\centi\metre}$, $d_o=8.6^\circ$).  Boxes show the
           interquartile range and mean (triangle); points are individual
           trials.  All trials fall below the tolerance line
           (dashed, $=1$).}
  \label{fig:hw_ee_error}
\end{figure}

Targets are drawn from the restricted task envelope of
Section~\ref{sec:workspace}: for each trial, a Cartesian target
$(\vect{p}^*,\mat{R}^*)$ is generated within the apple-picking bounding box
with $x\geq\SI{0.1}{\metre}$, position IK is solved (residual reported as
``IK residual'' below), and---only if the residual is below
\SI{2}{\centi\metre}---the trial proceeds; otherwise the target is
re-sampled.  This single check operationalises the reachability analysis of
Section~\ref{sec:workspace} for hardware use.  Three independent batches of
$10$ targets each ($n=30$ total) were executed.  Given the small sample size,
we report the achieved end-effector position and orientation error for every
trial rather than a binary success rate: the position error is
$2.64\pm1.04$\,cm (worst case $3.94$\,cm) and the orientation error is
$6.92\pm1.33^\circ$ (worst case $8.58^\circ$), i.e., every one of the $30$
trials lands within the $\SI{4}{\centi\metre}/8.6^\circ$ tolerance used during
training, with a margin of at least $\SI{0.06}{\centi\metre}$ even in the
worst-case trial (Fig.~\ref{fig:success_funnel}b,
Fig.~\ref{fig:hw_ee_error}).  Joint angles, velocities, and commanded torques
are logged at \SI{100}{\hertz} and passed through the identified power
model~\eqref{eq:power} to obtain the energy figures below, exactly as for the
simulation evaluation of Section~\ref{sec:sim_results}.

\begin{table}[t]
  \caption{Real-robot validation, $n=30$ (3 batches $\times$ 10 targets); all
           trials land within the $\SI{4}{\centi\metre}/8.6^\circ$ tolerance
           (Fig.~\ref{fig:success_funnel}b).  ``Exc.''\ is the normalised
           joint excursion $|q-\bar q|/(\Delta q/2)$.}
  \label{tab:hw_results}
  \centering
  \begin{tabular}{lccc}
    \toprule
    Metric & Mean $\pm$ std & Median & Range \\
    \midrule
    Pos.\ error (cm)            & $2.64 \pm 1.04$  & $2.77$ & $[0.70, 3.94]$ \\
    Ori.\ error ($^\circ$)      & $6.92 \pm 1.33$  & $7.16$ & $[3.79, 8.58]$ \\
    IK residual (cm)            & $0.69 \pm 0.45$  & $0.59$ & $[0.14, 1.49]$ \\
    Energy (J)                  & $71.5 \pm 48.3$  & $54.9$ & $[11.6, 165.2]$ \\
    Duration (s)                & $0.545 \pm 0.381$ & $0.374$ & $[0.10, 1.31]$ \\
    Mean power (W)              & $131.5 \pm 18.1$ & $130.7$ & $[95.0, 162.1]$ \\
    Mean $|\dot q|$ (rad/s)     & $1.85 \pm 0.27$  & $1.87$ & $[1.20, 2.47]$ \\
    Max $|\dot q|$ (rad/s)      & $6.26 \pm 1.27$  & $6.28$ & up to $8.79$ \\
    Mean $|\tau|$ (N\,m)        & $3.99 \pm 0.41$  & $4.01$ & $[2.91, 4.94]$ \\
    Max $|\tau|$ (N\,m)         & $15.2 \pm 3.1$   & $15.7$ & up to $20.9$ \\
    Steps per episode           & $45.2 \pm 30.4$  & $33$   & $[8, 105]$ \\
    Mean excursion              & $0.157 \pm 0.068$ & $0.186$ & $[0.05, 0.24]$ \\
    \bottomrule
  \end{tabular}
\end{table}

\begin{table*}[t]
  \caption{Real-robot validation, $n=30$: per-joint kinematics/torques (mean
           over all episodes, max over all steps) and post-hoc energy
           breakdown~\eqref{eq:power} (mean over all episodes).}
  \label{tab:hw_perjoint}
  \centering
  \begin{tabular}{lcccccc}
    \toprule
    Joint & $|\dot q|$ mean (rad/s) & $|\dot q|$ max & $|\tau|$ mean (N\,m) &
    $|\tau|$ max (N\,m) & Exc.\ mean & Exc.\ max \\
    \midrule
    Sh.\ pitch & 1.46 & 5.45 & 7.69 & 20.25 & 0.119 & 0.626 \\
    Sh.\ roll  & 1.63 & 7.19 & 6.03 & 20.88 & 0.122 & 0.534 \\
    Sh.\ yaw   & 2.41 & 7.60 & 4.04 & 17.06 & 0.094 & 0.286 \\
    Elbow      & 1.78 & 6.60 & 5.36 & 15.50 & 0.193 & 0.597 \\
    Wr.\ roll  & 2.11 & 8.79 & 1.16 & 14.69 & 0.097 & 0.503 \\
    Wr.\ pitch & 1.77 & 8.59 & 2.14 &  6.96 & 0.137 & 0.554 \\
    Wr.\ yaw   & 1.80 & 6.15 & 1.53 &  6.95 & 0.148 & 0.519 \\
    \bottomrule
  \end{tabular}
  \vspace{4pt}

  \begin{tabular}{lccccc}
    \toprule
    Energy term & Mechanical & Copper & Coulomb & Viscous & Interaction \\
    \midrule
    Share (\%) & $17.5 \pm 5.6$ & $46.1 \pm 8.4$ & $2.1 \pm 0.3$ & $20.4 \pm 4.3$ & $13.9 \pm 3.5$ \\
    \bottomrule
  \end{tabular}
\end{table*}

\begin{figure}[t]
  \centering
  \includegraphics[width=\columnwidth]{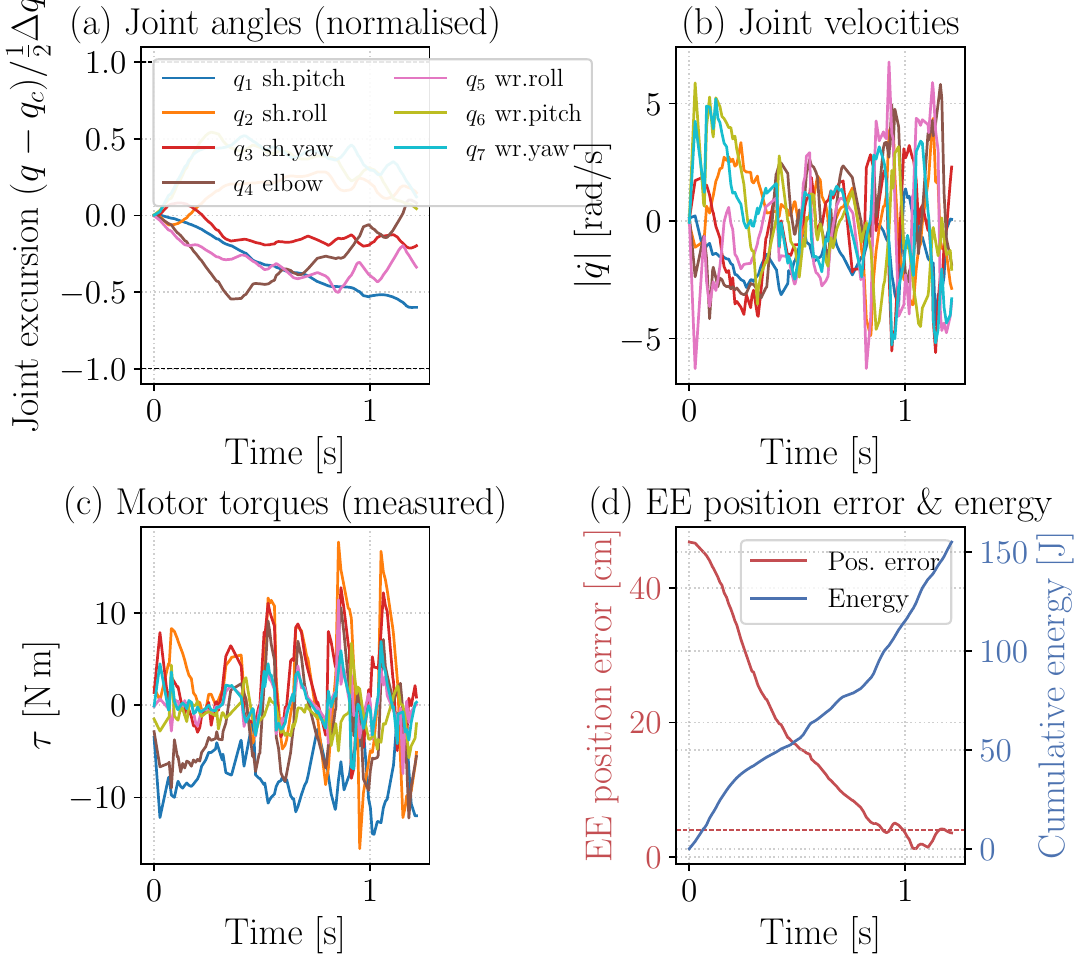}
  \caption{Representative hardware trial ($101$ steps, $\sim$\SI{1}{\second}).
           (a)~Per-joint excursion $(q-\bar q)/(\Delta q/2)$, with $\pm 1$
           marking the joint limits.  (b)~Joint velocities.  (c)~Measured
           motor torques.  (d)~End-effector position error (left axis) and
           cumulative energy via the power model~\eqref{eq:power} (right
           axis).}
  \label{fig:hw_trajectory}
\end{figure}

\begin{figure}[t]
  \centering
  \includegraphics[width=\columnwidth]{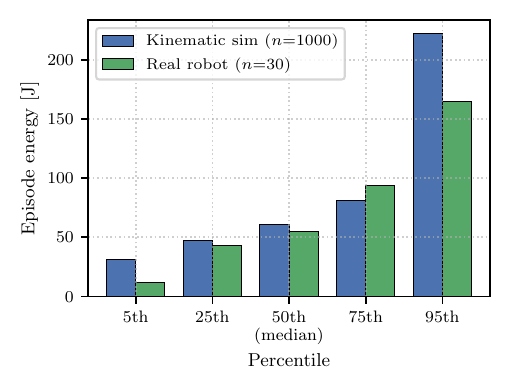}
  \caption{Episode-energy percentiles: kinematic simulation ($n{=}1000$, all
           episodes) vs.\ real hardware ($n{=}30$, all successful).  The
           hardware distribution is both lower and narrower than the
           kinematic-simulation distribution at every percentile, consistent
           with the lower torques and shorter episodes (Table~\ref{tab:hw_results})
           observed under the restricted task envelope of
           Section~\ref{sec:workspace}.}
  \label{fig:energy_distribution}
\end{figure}

Across the $n=30$ hardware trials, energy consumption (mean $71.5\pm48.3$\,J,
median $54.9$\,J) is markedly lower than the kinematic-simulation mean for
successful episodes ($98.16$\,J, Table~\ref{tab:sim_results}), and its
distribution is shifted toward lower values at every percentile
(Fig.~\ref{fig:energy_distribution}), reflecting the smaller realised torques
(up to $\SI{20.9}{\newton\metre}$, Table~\ref{tab:hw_perjoint}, vs.\ up to
$\SI{199}{\newton\metre}$ in kinematic simulation,
Table~\ref{tab:sim_perjoint}) imposed by the soft PD controller and torque
clamp of Section~\ref{sec:mujoco_validation}.  Correspondingly, the energy
breakdown shifts away from copper-dominated: copper losses fall from
$87.6\%$ (kinematic sim) to $46.1\pm8.4\%$ on hardware, with viscous friction
($20.4\pm4.3\%$) and mechanical work ($17.5\pm5.6\%$) accounting for a larger
share.  Final orientation error (mean $6.92\pm1.33^\circ$,
Fig.~\ref{fig:hw_ee_error}) sits close to, but consistently within, the
$8.6^\circ$ tolerance---the orientation channel is the tighter of the two
success criteria on hardware, consistent with the MuJoCo finding
(Section~\ref{sec:mujoco_validation}) that orientation tracking carries most
of the residual sim-to-real error.

% =============================================================================
\section{Discussion}
\label{sec:discussion}

\subsubsection*{Reward design: energy proxy vs.\ post-hoc evaluation}

Section~\ref{sec:reward} deliberately separates the quantity that
\emph{shapes} the policy from the quantity used to \emph{evaluate} it.  The
torque-norm proxy $\lambda_\tau\norm{\vect{\tau}_t}^2$ is cheap, smooth, and
correlates with the copper-loss term of~\eqref{eq:power} (both scale with
$\tau^2$), but it omits the viscous, Coulomb, and interaction terms and the
sign-dependent mechanical-work term.  With $\alpha=0$, the full power
model~\eqref{eq:power} never enters the gradient; it is computed purely for
post-hoc evaluation (Remark~\ref{rem:energy_proxy}).  A notable consequence,
visible in Sections~\ref{sec:sim_results} and~\ref{sec:hardware}, is that the
\emph{realised} energy reduction on hardware (median $54.9$\,J vs.\ a
kinematic-simulation mean of $98.16$\,J on successful episodes) is driven less
by the reward's energy term than by the deployment PD-gain choice
($k_p^{\mathrm{mj}}=50$) and the restricted, lower-effort task envelope of
Section~\ref{sec:workspace}.  This suggests that enabling $\alpha>0$, or
increasing $\lambda_\tau$, would act as an additional, currently unused lever
for further energy reduction, at the cost of a fresh training run
(Section~\ref{sec:lineage}).

\subsubsection*{Sim-to-real gap: mechanism}

The kinematic environment of Section~\ref{sec:mdp} assumes that the commanded
velocity is realised exactly within one step.  The MuJoCo PD-gain sweep
(Table~\ref{tab:kp_sweep}) shows that this assumption is the dominant source of
the sim-to-real gap: the success-rate drop ($20$--$25$ percentage points) and
the orientation-tracking inflation ($11$--$13^\circ$ above the $8.6^\circ$
training tolerance) are both largely \emph{independent} of the PD gain, which
is the signature of a structural mismatch between the training assumption and
any finite-bandwidth controller, rather than a tuning deficiency.  Unlike the
generic mitigations suggested for this class of gap---domain randomisation of
inertial parameters, or fine-tuning on real-robot data---the PD-gain sweep
allowed us to \emph{quantify} the gap and choose deployment gains
($k_p^{\mathrm{mj}}=50$, $k_d^{\mathrm{mj}}=1.5$) that keep saturation and
energy near the low end of the sweep (Fig.~\ref{fig:kp_sweep}) without
additional training.

\subsubsection*{Workspace reachability as the dominant factor in the success gap}

A second, largely independent contributor to the gap between the $69.9\%$
kinematic rate and the $46\%$ full-workspace MuJoCo rate is geometric, not
dynamic: the kinematic environment samples targets in \emph{joint space}
(Section~\ref{sec:target_sampling}), which guarantees reachability by
construction, whereas a deployment-style sampling of Cartesian targets within
the nominal apple-picking bounding box is reachable in only $\sim30\%$ of
cases (Table~\ref{tab:reachability}).  The remaining $\sim70\%$ miss by a
median of $\sim\SI{10}{\centi\metre}$---an order of magnitude larger than the
\SI{4}{\centi\metre} tolerance---which no controller, however well tuned, can
close.  Once this geometric mismatch is removed by restricting to the
reachable task envelope, the MuJoCo screening recovers a $95\%$ success rate
($n=20$), at or above the kinematic reference; on the real robot ($n=30$),
every trial lands within tolerance, with end-effector position errors
(Fig.~\ref{fig:success_funnel}b) comparable to or tighter than the kinematic
reference.  This indicates that, for a deployed system, target
selection (or a perception/grasp-planning stage that proposes only reachable
grasp poses) is at least as important as policy or controller tuning for
overall task success.

\subsubsection*{Energy: simulation vs.\ hardware}

The energy breakdown shifts substantially between kinematic simulation and
hardware (Tables~\ref{tab:sim_perjoint} and~\ref{tab:hw_perjoint}).  In
simulation, copper losses dominate ($87.6\%$), driven by RNEA torques up to
$\SI{199}{\newton\metre}$ on the shoulder-pitch joint---torques that arise
from the large accelerations a perfect-tracker policy can command.  On
hardware, the soft PD controller and torque clamp
(Section~\ref{sec:mujoco_validation}) limit realised torques to at most
$\SI{20.9}{\newton\metre}$, and the energy mix becomes more balanced (copper
$46.1\%$, viscous $20.4\%$, mechanical $17.5\%$, interaction $13.9\%$,
Coulomb $2.1\%$).  The identified power model~\eqref{eq:power} correctly
captures both regimes because all five terms remain active; only their
relative shares change with the torque/velocity operating point, which is a
useful sanity check on the model's generality beyond the identification data.

\subsubsection*{Orientation tracking}

On hardware, the mean final orientation error ($6.92\pm1.33^\circ$,
Fig.~\ref{fig:hw_ee_error}) sits close to, but consistently below, the
$8.6^\circ$ tolerance, mirroring the MuJoCo finding that orientation tracking
absorbs most of the residual sim-to-real error
(Section~\ref{sec:mujoco_validation}).  The Hybrid Constellation Reward
(Section~\ref{sec:reward}) couples position and orientation into a single
geometric quantity $d_{\mathrm{con}}^t$, which removes the need to separately
tune a position/orientation weight ratio (a known difficulty in the earlier,
separable reward, Section~\ref{sec:lineage}); nonetheless, orientation remains
the tighter of the two success criteria in practice, and is the most likely
margin to be lost if the task envelope were widened further.

\subsubsection*{Limitations}

The kinematic-simulation result ($69.9\%$, $n=1\,000$, fixed seed) follows the
$n\ge1000$ protocol adopted after the training lineage of
Section~\ref{sec:lineage} showed that smaller-$n$ improvements did not
reproduce at scale.  The MuJoCo restricted-envelope screening ($n=20$) and the
real-robot validation ($n=30$) are, by contrast, feasibility-level
demonstrations on a single fixed target set within the reachable task
envelope; they establish that the policy \emph{can} operate reliably in this
envelope, but do not constitute a formal $n\ge1000$ success-rate estimate for
hardware.  Likewise, the reward's torque-norm energy proxy
(Remark~\ref{rem:energy_proxy}) was not directly compared, at training scale,
against training with $\alpha>0$ using the full power model; the lineage of
Section~\ref{sec:lineage} indicates that such changes require a fresh
$5\times10^6$-step run and were outside the project's remaining compute
budget.

% =============================================================================
\section{Conclusion}
\label{sec:conclusion}

This paper presented an end-to-end, energy-aware reinforcement learning
framework for the 7-DOF left arm of the Unitree~G1 humanoid, validated along a
four-stage arc from kinematic simulation to physical hardware.  An
experimentally identified electrical power model
($R^2=0.933$, RMSE$=\SI{1.07}{\watt}$, hold-out $R^2=0.965$) decomposes
per-joint power into mechanical, copper, Coulomb, and viscous terms plus
pairwise interactions, and serves both as a training-time energy proxy
($\lambda_\tau\norm{\vect{\tau}}^2$) and as the common evaluation metric for
every experiment in this paper.  A SAC policy combining an incremental
joint-position action space with a Hybrid Constellation Reward reaches
$69.9\%$ success over $n=1\,000$ kinematic-simulation targets.  A frozen-policy
MuJoCo validation with a PD-gain sweep showed that this rate drops to $46\%$
under realistic actuator dynamics ($n=200$), with an intrinsic
$11$--$13^\circ$ orientation-tracking penalty arising from the kinematic
environment's perfect-tracker assumption.  A workspace-reachability analysis
attributed most of the remaining gap to geometry rather than dynamics: only
$\sim30\%$ of a nominal apple-picking bounding box is reachable within
\SI{2}{\centi\metre}.  Restricting evaluation to this reachable task envelope
recovered $95\%$ success in MuJoCo ($n=20$) and, on the physical Unitree~G1
over three independent batches ($n=30$), end-effector position/orientation
errors of $2.64\pm1.04$\,cm / $6.92\pm1.33^\circ$ (worst case $3.94$\,cm /
$8.58^\circ$)---inside the $\SI{4}{\centi\metre}/8.6^\circ$ training tolerance
in every trial---at a median energy of $\SI{54.9}{\joule}$.

We emphasise that this work constitutes a \emph{first approach} in the context
of robotic apple harvesting using humanoid robots.  The reaching task studied
here---moving the end-effector to a target pose within an energy budget---is
the core repeated motion of an autonomous picking cycle, but a deployed
harvesting system additionally requires perception and fruit localisation,
grasp planning and execution, and coordination with locomotion or torso
positioning to bring the target within the reachable task envelope identified
in Section~\ref{sec:workspace}.  The validation arc presented here---moving
from a kinematic training assumption, through a dynamics-level sim-to-real
gap analysis, to a geometrically restricted but physically validated task
envelope---is intended as a template for how such energy-aware arm policies
can be progressively hardened toward field deployment on a harvesting
platform, rather than as a complete harvesting system.

Future work will focus on: (i)~training with $\alpha>0$ (full power-model
reward) from the current checkpoint, now that the torque-proxy formulation has
established a working baseline; (ii)~expanding the reachable task envelope
through torso or base repositioning, so that a larger fraction of the nominal
workspace becomes usable without violating the reachability constraints of
Section~\ref{sec:workspace}; (iii)~integrating fruit-detection and grasp-pose
estimation so that target poses are generated from perception rather than
sampled synthetically; and (iv)~scaling the real-robot validation toward the
$n\ge1000$ protocol used in simulation, as compute and operator time allow.

% =============================================================================
\appendix
\section{Minimum-Jerk Trajectory Derivation}
\label{app:minjerk}

Given start position $x_0 \in \R^m$ with zero velocity and acceleration, and
a target $x^*$ to be reached at time $T$, the minimum-jerk trajectory
minimises $\int_0^T \norm{\dddot{x}}^2\,dt$ and has the closed form
\cite{Flash1985}:
\begin{equation}
  x(t) = x_0 + (x^* - x_0)
  \left[10\xi^3 - 15\xi^4 + 6\xi^5\right],
  \quad \xi = t/T.
  \label{eq:minjerk}
\end{equation}
For evaluation, $x$ is replaced by the joint configuration vector
$\vect{q} \in \R^7$ and the duration is set automatically as
$T = (15/8)\,\Delta q_{\max}/\dot{q}_{\max}$, where
$\Delta q_{\max} = \max_i |q^*_i - q_{0,i}|$, so that the maximum joint
velocity never exceeds $\dot{q}_{\max} = \SI{2.5}{\radian\per\second}$.
This gives a trajectory duration of $1.519 \pm 0.342$\,s on average across the
target-sampling distribution of Section~\ref{sec:target_sampling}, compared to
the RL episode budget of \SI{3.0}{\second}.

% =============================================================================
\bibliographystyle{IEEEtran}
\bibliography{biblio}

\end{document}